\documentclass[conference]{IEEEtran}
\IEEEoverridecommandlockouts
\usepackage{cite}
\usepackage[switch,columnwise]{lineno}
\usepackage{amsmath,amssymb,amsfonts}
\usepackage{algorithm}
\usepackage{amsthm}
\newtheorem{defn}{Definition}
\usepackage{algpseudocode}
\usepackage{graphicx}
\usepackage{textcomp}
\usepackage{xcolor}
\usepackage{float}
\usepackage{url}
\usepackage{booktabs}
\usepackage[numbers]{natbib}
\def\BibTeX{{\rm B\kern-.05em{\sc i\kern-.025em b}\kern-.08em
    T\kern-.1667em\lower.7ex\hbox{E}\kern-.125emX}}
\begin{document}

\title{A Utility-Mining-Driven Active Learning Approach for Analyzing Clickstream Sequences\thanks{This paper is an extended version of the work available on [arXiv:2410.07282].}\\}

\author{
    \IEEEauthorblockN{Danny Y. C. Wang, Lars Arne Jordanger, and Jerry Chun-Wei Lin\\}
    \IEEEauthorblockA{\textit{Department of Computer Science, Electrical Engineering and Mathematical Sciences} \\
    \textit{Western Norway University of Applied Sciences}\\
    Bergen, Norway \\
    \{ycw, Lars.Arne.Jordanger, jerry.chun-wei.lin\}@hvl.no} 
}

\maketitle

\begin{abstract}
As e-commerce evolves rapidly, the capability of selecting informative data for model training is essential. This selective data sampling approach aims to minimize manual labeling costs while maintaining or improving model prediction accuracy. However, most existing active learning models lack sufficient explainability in their selection process. To address this limitation, this study proposes HUSPM-SHAP, a utility mining-based active learning method that uses SHAP values to identify the most informative clickstream sequences, improving both interpretability and model updates. Our experiments show that careful tuning of SHAP parameters for positive and negative classes is critical, as it directly impacts the selection outcomes. Extensive results on purchase prediction highlight that HUSPM-SHAP reliably reduces labeling effort while maintaining strong predictive accuracy across scenarios. To the best of our knowledge, this represents the first study to employ the concept of high-utility mining to augment active learning processes, marking a pioneering approach in the field. Our findings demonstrate the model's capability to refine e-commerce data processing, steering towards more streamlined, cost-effective prediction modeling.
\end{abstract}


\section{Introduction}\label{introduction}

Clickstream data records user actions and the sequence of these actions, providing detailed insights into customer interactions and decision-making processes \cite{requena2020shopper}. It not only tracks user actions on a website but also captures the sequence of these actions, enabling a detailed view of customer interactions and decision-making processes. The outcomes from sequence-based analysis provide an in-depth understanding of user behavior, facilitating enhancements in the online shopping experience and achieving business objectives.

One interesting aspect of clickstream data analysis is to understand the behavior of visitors who browse a site but do not add items to their cart or complete a purchase. Analyzing this particular group can shed light on potential obstacles in the user's shopping behavior. Such analysis helps identify areas for website improvement, including navigation, the checkout process, and product information accessibility, supporting increased user engagement and conversion rates.


E-commerce clickstream data analysis has developed quickly, moving beyond simply describing user behavior to adopting predictive methods \cite{requena2020shopper, awalkar2016prediction, ozyurt2022deep, rausch2022predicting, toth2017predicting}. These approaches, transitioning from simply understanding to proactively predict user actions, signify a substantial step forward in utilizing clickstream data to better e-commerce strategies. This shift represents an advancement in the realm of e-commerce, enhancing the ability to engage customers effectively and improving business strategies. A previous study \cite{7539341} identifies four levels of granularity in clickstream analysis including patterns, segments, sequences and events. Sequential pattern is indicative of navigation loops through repeated events in a sequence \cite{7539341}, signaling potential customer behaviors and sales opportunities.

\begin{figure}[!t]
  \centering
  \includegraphics[width=\linewidth]{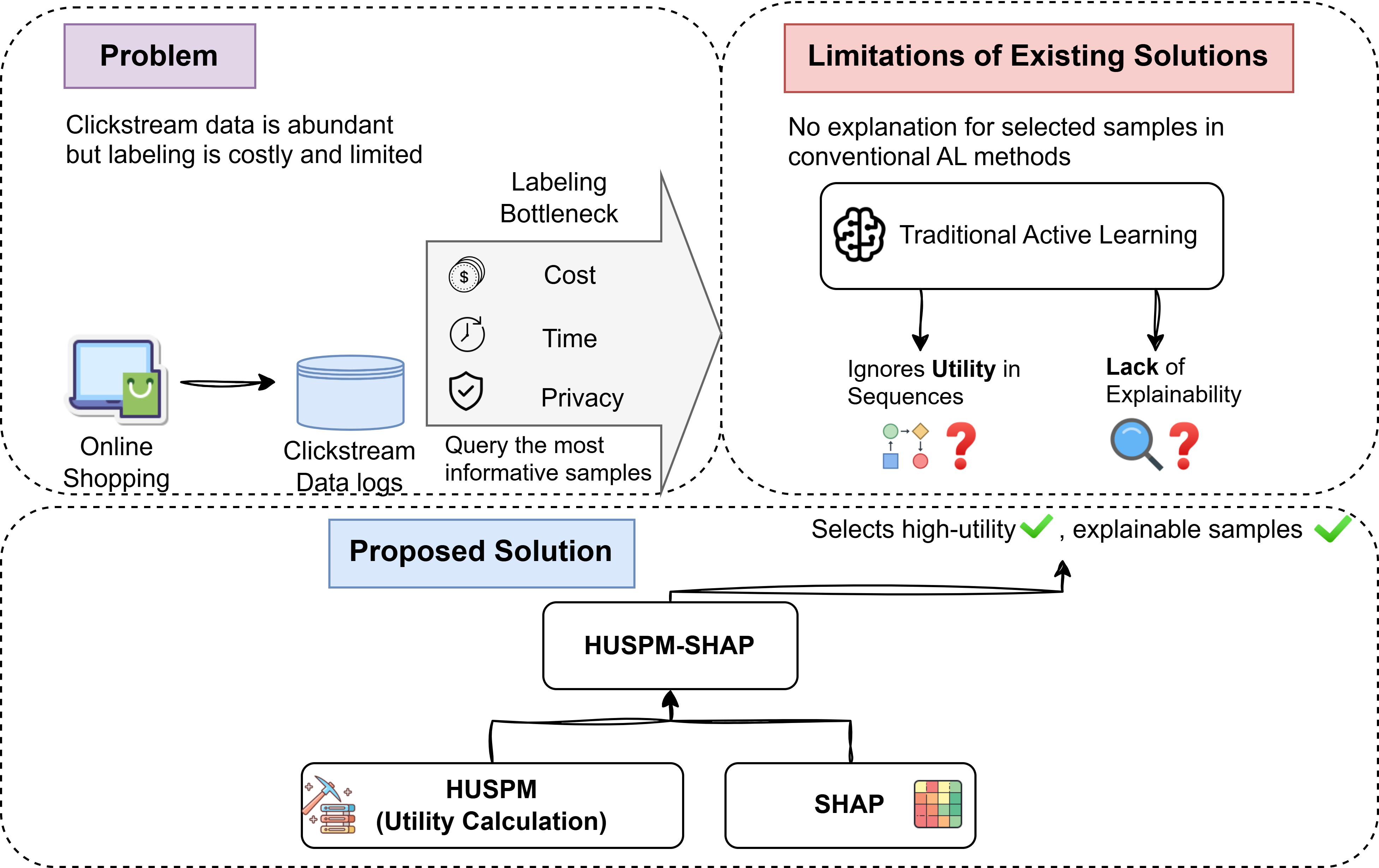}
  \caption{Motivational overview illustrating the problem and limitations of existing active learning methods.}
  \label{fig:motivation}
\end{figure}

Dealing with the labeling and processing of large volumes of data presents a key challenge in the context of e-commerce clickstream data. The rapid generation of data on e-commerce platforms often leads to large volumes of data \cite{pal2018big}, an important portion of which remains neglected due to data pre-processing and labeling procedures falling behind. The accumulation of unhandled data implies much of it goes unused, reducing training efficiency and making it more difficult for models to adapt to user behavior changing. Confidentiality concerns in e-commerce add to the difficulty \cite{srivastava2023commerce}, as outsourcing data labeling is often not feasible owing to the need to protect sensitive business information. Clickstream datasets themselves can be extremely large, with the number of sessions ranging from 2.7 to 37 million interactions \cite{tagliabue2021sigir}, demonstrating the substantial volumn of user interations these platforms manage. Furthermore, the complexity of customer behaviors, such as varied purchasing intents and shopping cart abandonment, requires considerable effort and resources for accurate labeling. The challenge is further intensified by the often imbalanced datasets used for predicting shopping intent in clickstream data \cite{requena2020shopper, ozyurt2022deep}.

Active Learning (AL) addresses the high labeling cost in clickstream data by selecting the most informative samples \cite{mosqueira2023human}. However, as illustrated in Fig.~\ref{fig:motivation}, existing AL methods lack explainability and ignore the utility of user behavior sequences. To address these limitations, we propose HUSPM-SHAP, which integrates utility mining with SHAP \cite{lundberg2017unified} to improve sample quality and interpretability.

Recently, Kara et al. \cite{kara2022shap} proposed a novel active learning strategy designed for text classification. This method is applicable in banking domains, where privacy concerns for data labeling are limited. This method prioritizes challenging instances for human annotation based on SHAP values \cite{lundberg2017unified}, enhancing training sets iteratively. When it applies to the classification of customer comments in banking surveys, this approach surpasses two existing strategies, particularly in scenarios with class imbalances, showcasing the effectiveness of integrating human judgment in active learning for quality dataset creation. However, this study focuses merely on filtering single elements, such as individual words within a sentence, instead of extracting several sequences. Whether in text classification or clickstream of early prediction, the data sources primarily comprises sequences; understanding this characteristic is essential for effective data selection, leading to more precise analyses and results.

\subsubsection{Research Gap}

Although current active learning approaches have demonstrated effectiveness in reducing labeling workload, they often lack sufficient transparency, failing to intuitively explain why specific data points are selected for labeling. For example, Kara et al. \cite{kara2022shap} pioneered a pool-based active learning strategy that employs SHAP values for text classification. Specifically, their method focuses on sequentially selecting elements based on the highest SHAP values, prioritizing individual words. However, this approach does not extend to assessing the overall utility of data sequences, an area yet to be explored in e-commerce clickstream data analysis. This study seeks to address this limitation by adopting utility mining techniques, seeking to enhance model transparency. It does so by systematically identify and prioritize valuable sequences based on their utility, hence filling the gap.

\subsubsection{Research Objective}
 This study addresses these limitations by introducing a novel active learning framework, HUSPM-SHAP, that integrates High-Utility Sequential Pattern Mining (HUSPM) with SHAP values to:
\begin{itemize}
\item Identify and prioritize the most informative sequences in a pool-based active learning setting. By focusing on these sequences, the framework ensures that the labeled data selected for model training contributes significantly to performance improvements.
\item Enhance the transparency and interpretability of the data selection process. This integration allows for clear insights into why certain sequences are chosen, supporting better understanding and decision-making in active learning workflows.
\item Support prediction performance by prioritizing high-utility sequences from the unlabeled data pool during training. By leveraging these sequences, the framework aims to improve the model’s ability to predict user behavior patterns effectively, even in imbalanced scenarios.
\end{itemize}

This framework is applied to e-commerce clickstream data to address challenges such as the high cost of labeling. Experiments demonstrate that HUSPM-SHAP can reduce labeling requirements while maintaining or improving prediction performance. By leveraging the strengths of both HUSPM and SHAP, this approach offers a solution for discovering valuable sequences and achieving efficient active learning outcomes.

\section{Related Work}
\subsection{Prediction Models of Clickstream E‑commerce Data}\label{EarlyPrediction}

Recently, clickstream data analysis in e-commerce primarily focused on understanding behavioral patterns rather than prediction. However, predictive approaches have only recently started emerging in this field \cite{ulitzsch2023machine}. Building upon such methodologies, some studies specifically aim to predict shopper intentions in e-commerce using early-window clickstream data \cite{requena2020shopper, ulitzsch2023machine}. 

In the context of this study, the conceptualization of early prediction within e-commerce clickstream data analysis is grounded in the framework. Early prediction refers to the ability to predict a user's future actions, such as making a purchase, by analyzing only the initial part of their activity sequence on a website. For example, Requena et al. \cite{requena2020shopper} analyzed shopper intent by browsing sessions that includes action sequences ranging from $5$ to $155$ steps, which is typical for clickstream data in interactive tasks.

In e-commerce, a notable phenomenon is that customers often add items to their virtual shopping carts without immediate plans to purchase. This behavior, known as shopping cart abandonment, is an essential aspect of online shopping behavior. It occurs when consumers place items in their online cart but do not follow through with the purchase in the same session, highlighting a key area for analysis in e-commerce strategy \cite{rausch2022predicting}. Other studies involve the assessment of the likelihood of users exiting a site without completing purchases, a challenge commonly addressed in clickstream analysis \cite{requena2020shopper, awalkar2016prediction, ozyurt2022deep, toth2017predicting}. Early detection of potential purchase risks enables automated strategies, such as offering discounts, to encourage customers towards making a purchase. To facilitate this early detection, various supervised approaches have been utilized. These include methods such as hidden Markov models \cite{ozyurt2022deep} and recurrent neural networks \cite{toth2017predicting}, which are effective for analyzing sequential data.

\subsection{Active Learning}
Active Learning (AL) \cite{settles2009active}, a specialized branch of machine learning, emerges as a powerful tool in effectively analyzing vast and complex data \cite{mosqueira2023human}. Diverging from conventional machine learning techniques, which passively process existing datasets, AL proactively identifies and labels new data points, thereby continuously improving its learning algorithm. This method is particularly beneficial in situations where data is plentiful but labeled examples are rare or costly to acquire. 
It is essential to differentiate between several key processes within active learning. The foremost among these is the sampling process, also referred to as the query strategy \cite{mosqueira2023human}. This specific process involves the selection of instances that require labeling by a human expert. Munro \cite{munro2020rapid} outlines three key sampling strategies: random, uncertainty, and diversity. The random approach selects data for labeling by chance, which is a straightforward process. However, the uncertainty and diversity strategies present a more complex scenario, balancing the choice between exploiting known information and exploring new possibilities. Recent advances have introduced methods such as the Core-set approach, which focuses on maximizing coverage and diversity by selecting representative subsets of the data pool \cite{sener2017active}. Learning Loss for Active Learning (LLAL) \cite{yoo2019learning} uses an additional small neural network, called a meta-model. The meta-model predicts which unlabeled samples would be the most helpful for improving the main model if their labels were known \cite{yoo2019learning}. AL models can be categorized into two types. The first is Sequential Active Learning, and the latter one is Pool-based Active Learning \cite{settles2009active}. Sequential Active Learning is a process where a machine learning algorithm evaluates and selects individual data points one at a time from a stream of unlabeled data for labeling \cite{settles2009active}; pool-based Active Learning is a method where a machine learning algorithm selectively chooses the most informative data points from a fixed pool of unlabeled data for labeling \cite{settles2009active}.

\subsection{High-Utility Sequential Pattern Mining}

High-utility sequential pattern mining (HUSPM) represents a growing field in data mining that focuses on identifying high-utility sequential patterns (HUSPs) from quantitative databases, considering both the utility and the sequence of data elements \cite{9016187}. In the context of decision-making, adjusting the threshold to extract an optimal number of HUSPs can be time-consuming. The top-k HUSPM approach \cite{yin2013efficiently} simplifies this by focusing on the most relevant patterns, such as identifying the top-k profitable products in a retail setting \cite{zhang2021tkus}. The calculation example and tables can be found in Section~\ref{subsec:huspm_bg}.

\subsection{Explainability of SHAP Value}\label{explainability}
Shapley Additive Explanations (SHAP), grounded in cooperative game theory, offers a unified method to elucidate machine learning (ML) predictions by combining SHAP values and feature variances \cite{lundberg2017unified}. The significance of a specific feature within a model can be ascertained by averaging the absolute SHAP values across all instances, providing a comprehensive view of the feature's influence on the model's output. SHAP values assign a significance score to each feature for a specific prediction, demonstrating an additive property in determining feature significance \cite{lundberg2017unified}.

\subsubsection{Additive Feature Attribution Methods for ML Model Explanation}
 An essential attribute of machine learning model explanation is the additive property \cite{lundberg2017unified}. According to \cite{lundberg2017unified}, a given prediction model \(f\) to be explained and an input feature vector \(x\), the prediction outcome \(f(x)\) and \(g\) represents the explanation model. The property implies that the predictive output of a model can be decomposed into a foundational baseline value supplemented by the sum of contributions of individual features towards the overall prediction result. Focusing on local methods aimed at explaining a single prediction model \(f(x)\), derived from an individual input \(x\), as initially introduced by LIME \cite{lundberg2017unified}, \cite{ribeiro2016why}. The simplified inputs \(x'\) maps to the original inputs \(x\) by the help of local mapping functions \(h_x\), such that \(x = h_x(x')\), and we have \(g(x') = f(x)\). Local approaches aim to achieve \(g(z') \approx f(h_x(z'))\) for any \(z' \approx x'\) \cite{lundberg2017unified}.

\begin{defn}
Additive feature attribution methods have an explanation model that is a linear function of binary variables:
\begin{equation}\label{formula1}
g(z') = \phi_0 + \sum_{i=1}^{M} \phi_i \cdot z_i',
\end{equation}
where \( z' \in \{0, 1\}^M \), \( M \) is the number of simplified input features, and \( \phi_i \in \mathbb{R} \).
\end{defn}

 \(\phi_0\) is the baseline value which is the expected output when no features are contributing to the prediction. It is also generally the average of all predictions on the training data. \(M\) is the number of features and \(\phi_i\) is the contribution value of feature \(i\) to the prediction outcome \(f(x)\), considered as SHAP value. Further details on the calculation and application of \(\phi\) will be covered in the upcoming section. The additive property allows us to explain a model's output as the sum of the linear contributions of each input feature and a baseline value. In this study, we leverage the additive nature of SHAP values as a utility measure in our proposed model to guide the selection of unlabeled data for active learning.

\section{The Developed Model}
In this study, we introduce a model called HUSPM-SHAP, which focuses on pool-based active learning designed specifically for analyzing e-commerce clickstream data. This data, capturing user interactions on a website, offers valuable insights into customer behavior. The HUSPM-SHAP model adopts an advanced retrieval strategy to select queries for acquiring higher-quality new training data, aiming to enhance the overall learning process and improve the predictive accuracy in the context of e-commerce clickstream analysis, and utilizing HUSPM combined with SHAP values as indicators of data utility. Furthermore, we aim to assess and enhance the model's effectiveness by progressively improving machine learning training through batch processing. 

\begin{figure*}[!htbp]
		\centering
		\centerline{\includegraphics[scale=0.72]{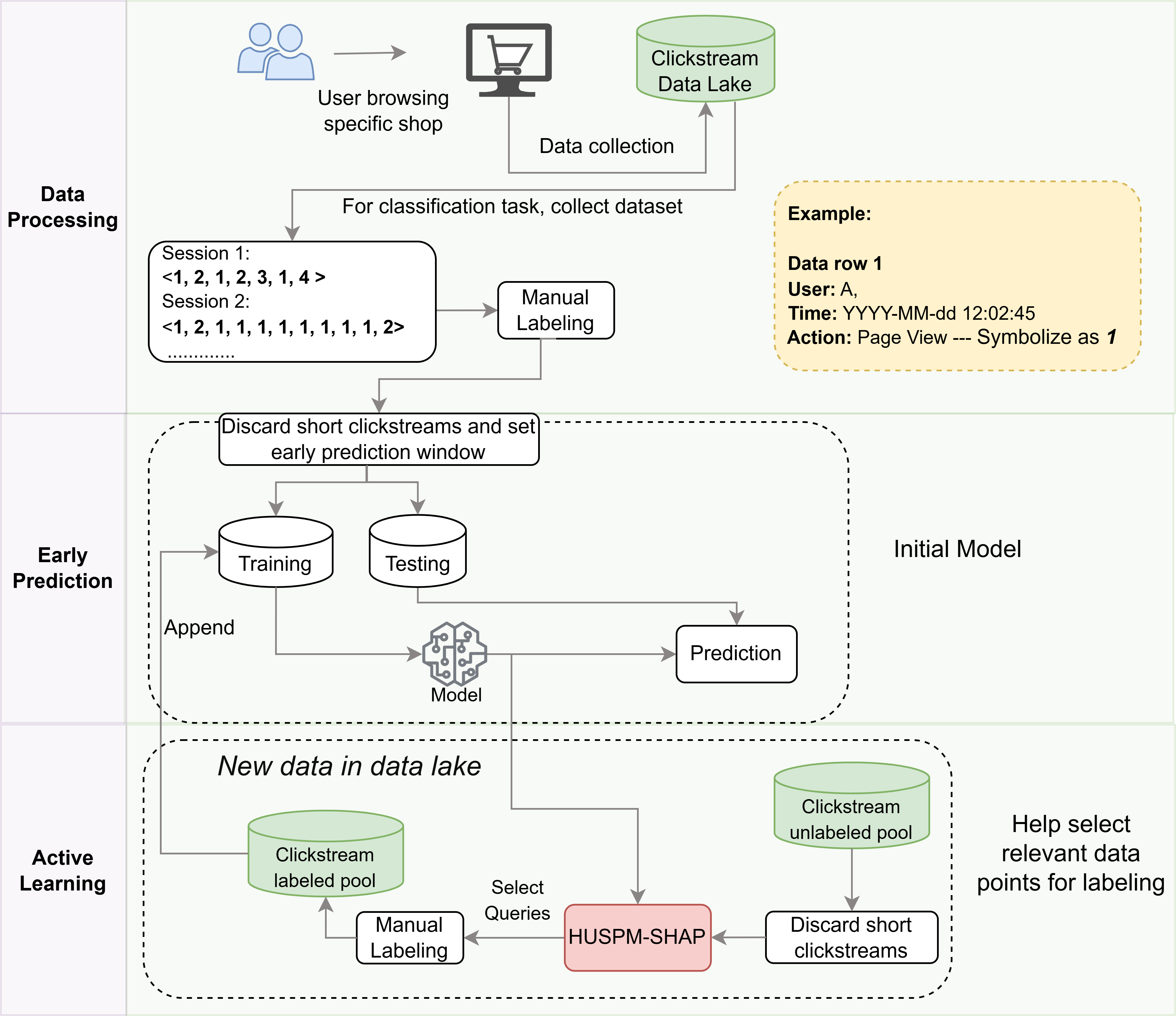}}
		\caption{A high-level overview of the clickstream analysis workflow employed in this study}
		\label{fig1}
	\end{figure*}

This method is designed to highlight both the efficiency and accuracy of the learning process by employing sophisticated analytical techniques and strategic data management. Fig. \ref{fig1} provides a high-level overview of the workflow employed in this study, and TABLE \ref{tab:my_label} shows symbolizing rules \cite{requena2020shopper} for clickstream data. In this case study, we will use the "Purchase" action as the label $\mathcal{Y}$ for our experiments. The variable
$\mathcal{Y}$ is defined as a binary indicator, where $\mathcal{Y} = 1$ represents a purchase occurrence, while $\mathcal{Y}=0$ indicates no purchase. This approach reflects real-world scenarios where labels are not readily available and must be derived from user interactions.

\begin{table}[!htbp]
\centering
\caption{Rules for assigning symbols to clickstream actions}
\begin{tabular}{lcc}
\hline
\textbf{Action} & \textbf{Symbol} & \textbf{Common Sequence Patterns} \\ \hline
Page view & 1 & \(\langle 1, 3 \rangle\), \(\langle 1, 2, 3 \rangle\) \\ 
Browsing product detail & 2 & \(\langle 2, 3 \rangle\) \\
Add product to shopping cart & 3 & \(\langle 1, 2, 3 \rangle\), \(\langle 1, 3 \rangle\), \(\langle 2, 3 \rangle\)\\
Remove product from cart & 4 & \(\langle 4, 1 \rangle\), \(\langle 4, 2 \rangle\) \\
\hline
\multicolumn{3}{l}{\textit{Note: The "Purchase" action is used as the label \( \mathcal{Y} \) in this study.}}
\label{tab:my_label}
\end{tabular}
\end{table}

\subsection{Problem Statement}

Given an initial e-commerce clickstream labeled dataset $\mathcal{D}_{\text{initial}}$ and an early prediction model $\mathcal{M}$ that has been trained on $\mathcal{D}_{\text{initial}}$ for a specific sequence length. Additionally, there is another unlabeled data pool $\mathcal{D}_{\text{unlabeled}}$. The objective is to develop a strategy for efficiently selecting a subset of unlabeled data points from $\mathcal{D}_{\text{unlabeled}}$. To evaluate the effectiveness of the selected data points, the model $\mathcal{M}$ conducts a re-training process using both $\mathcal{D}_{\text{initial}}$ and the newly labeled subset. This allows us to assess whether the selected data contributes to improved predictive performance.

Let $\mathcal{D}_{\text{unlabeled}} = \{x_1, x_2, \ldots, x_d\}$ represent the set of unlabeled clickstream sequences, where each $x_i$ is a sequence of click symbols from TABLE \ref{tab:my_label}. The challenge is to strategically select a high-quality subset $\mathcal{S} \subseteq \mathcal{D}_{\text{unlabeled}}$ for labeling and re-training. The goal is that, once $\mathcal{M}$ is re-trained on $\mathcal{D}_{\text{initial}} \cup \mathcal{S}$, the model maximizes prediction accuracy and robustness for early predictions of specific customer behaviors. In this case study, we focus on browsing behavior leading to a purchase, where the label $y_i \in \{0, 1\}$ indicates no purchase ($0$) or purchase ($1$).

The re-training process is an iterative procedure. In each iteration, a fixed number of newly labeled data points is added to the training set. However, the selection of these data points is determined by the top-$k$ most influential sequential patterns identified using the HUSPM-SHAP framework. Here, $k$ represents the number of high-utility patterns extracted rather than the number of new labeled data points. The number of new labeled data points remains constant in each iteration to ensure comparability across experiments. In different applications, $k$ can be dynamically adjusted based on the size of $D_{\text{unlabeled}}$ to optimize data selection efficiency. Since this is an early prediction problem, it is assumed that each sequence has a given length especially for referring to the number of clicks. The fixed sequence length makes sure consistency in the model’s input, as early prediction typically involves forecasting future behavior from a given step, such as after a certain number of clicks.

\subsection{Background and Utility Calculation Example}\label{subsec:huspm_bg}
To facilitate understanding of HUSPM framework, we introduce a simple example for utility calculation. Given a sequence database, each transaction has up to four elements regarding \(\langle1, 2, 3, 4\rangle\). The objective of HUSPM is to identify sequences with high utility values. The profit value associated with each element in a sequence is defined in TABLE \ref{table:utility}, and the following example sequences are presented in TABLE \ref{table:sequences}.

\begin{table}[!htbp]
\centering
\begin{minipage}{.5\linewidth}
\caption{An example utility table}
\centering
\begin{tabular}{cc}
\hline
\textbf{Element} & \textbf{Utility Value} \\
\hline
1 & \$10 \\
2 & \$15 \\
3 & \$20 \\
4 & \$5 \\
\hline
\end{tabular}
\label{table:utility}
\end{minipage}%
\begin{minipage}{.5\linewidth}
\caption{An example sequence database}
\centering
\begin{tabular}{cc}
\hline
\textbf{Sequence ID} & \textbf{Sequence} \\
\hline
A & \(\langle 1, 2, 1, 1, 3\rangle\) \\
B & \(\langle1, 2, 1, 1, 2\rangle\) \\
C & \(\langle2, 3, 4, 1, 2\rangle\) \\
\hline
\end{tabular}
\label{table:sequences}
\end{minipage}
\end{table}

The utility value of specific patterns within these sequences is computed as following examples such as:
\begin{itemize}
    \item Utility of pattern $\langle 1, 2 \rangle$ in Sequence A: $u(1) + u(2) = \$10 + \$15 = \$25$.
    \item Utility of pattern $\langle 1, 2 \rangle$ in Sequence B: $u(1) + u(2) = \$10 + \$15 = \$25$.
    \item Utility of pattern $\langle 2, 3 \rangle$ in Sequence C: $u(2) + u(3) = \$15 + \$20 = \$35$.
\end{itemize}
A utility threshold is set, for instance, $\theta = \$30$. A pattern is identified as a HUSP if its utility value exceeds this threshold. In this scenario, the pattern $\langle 2, 3 \rangle$ is considered a HUSP. It is important to note that in sequence B, the utility value of the pattern $\langle 1, 2 \rangle$ is computed as $25$, considering unique counting where repetitive patterns in the same transaction are counted only once or the highest utility is taken to prevent the overestimation of utility values for repeated patterns within a single transaction.

\subsection{HUSPM-SHAP model}
The HUSPM-SHAP introduces a systematic active learning strategy designed to identify high-utility unlabeled data within e-commerce clickstream analysis. The process begins with data preparation, where the dataset \( D \) is standardized and labeled for further analysis. Subsequently, a machine learning model \( f \) is developed, tailored specifically to \( D \) and trained to detect various user behaviors. A crucial component of this algorithm is the computation of SHAP values \cite{lundberg2017unified}, which elucidates the influence of each feature in \( D \) on the model's prediction.
We extend this analysis to examine the aggregated SHAP values for specific sub-sequences \( T \), consisting of distinct elements within the larger sequences \( S \). In this context, \( S \) represents a collection of sequences that include the element \( t \), with \( T \) denoting specific sub-sequences that capture meaningful patterns or interactions in the sequential data. This analysis involves calculating SHAP values for individual data instances using the Python SHAP library\footnote{\url{https://github.com/shap/shap/blob/master/docs/index.rst}} and aggregating these values across all appearances of sub-sequences \( T \) in \( S \). Each SHAP value \( \phi_i \) quantifies the marginal contribution of a feature \( i \) within a data instance to the model's output \( f \). \(\Phi_T\) for sub-sequences are computed by summing the contributions of each element or sub-sequence across all occurrences in \(S\) where they appear, thus providing a comprehensive understanding of the broader impact of specific features or behaviors. Detailed steps of the operations are given in Algorithm 1, and an intuitive example is given in Appendix.

\begin{enumerate}
\item \textbf{For an individual element \( t \)}:

Here, we consider each element to be a click. For an individual element \(t\), as shown in TABLE \ref{tab:my_label}, where \(t\) takes on values from the set \(\{1, 2, 3, 4\}\) representing actions within the clickstream scenario. Then we compute the aggregated SHAP values. These values capture the contribution of each action to the overall model prediction and are defined as follows:

\begin{equation}\label{aggregated1}
\Phi_t = \sum_{\ell} \max_{j} \{\phi_{\ell:tj}\},
\end{equation}

where \(\ell\) denotes the index of each data point, with \( \ell \in \{1, 2, 3, \ldots, d\} \). \(t\) represents the specific element type, and \(j\) indicates the position within the sequence. Instead of summing the SHAP values across all occurrences of \( t \) within a sequence, we take the maximum value to prevent overestimation of frequently occurring elements. This setup quantifies how the most influential occurrence of element \(t\) within a sequence affects its SHAP value, enabling us to measure the positional impact on feature importance more accurately.
\item \textbf{For a sub-sequence of elements \( T \)}:

Building upon the contribution of individual elements, we can extend this to evaluate the impact of a sub-sequence of consecutive and distinct elements \( T = \langle t_1, t_2, \ldots, t_n \rangle \), where \(t_{i+1} \neq t_i\). Here, \( T \) consists of values from the set \(\{1, 2, 3, 4\}\), representing consecutive behaviors combined by several actions without repetition. An \(n=2\) example could e.g. be the behavior of a customer transitioning from a page view to browsing product details as the subset \(T=\langle 1, 2 \rangle\). The formula for the general case for a sub-sequence \(T\) of length \(n\) is given by:

\begin{equation}\label{aggregated2}
\Phi_T = \sum_{\ell} \max_{q \in Q(\ell)} \{\sum_{k=1}^{n} \phi_{\ell:(q+k-1)}\},
\end{equation}

where \( Q(\ell) = \{q_1, q_2, \ldots, q_r\} \) represents the set of all starting positions of the sub-sequence \( T \) within the data point in \( S \) with index \( \ell \), \( q \) denotes the starting position of the sub-sequence \( T \) in the main sequence \( S \), and \( k \) is the index of each element within the sub-sequence \( T \), where \( k = 1, 2, \ldots, n \) with \( n \) being the length of \( T \). Instead of summing the SHAP values for all occurrences of \( T \) in a sequence, we take the maximum SHAP value sum among all its occurrences within the same sequence. The notation \(\phi_{\ell:(q+k-1)}\) captures the SHAP value of the element at position \( q+k-1 \) in the main sequence, corresponding to the position of the sub-sequence \( T \). This ensures that within each sequence, only the most influential occurrence of \( T \) is considered, preventing overestimation due to repeated patterns.

This approach allows us to assess the collective influence of sub-sequences \( T \) on the model’s predictions by considering the aggregated contributions of all elements based on their specific positions within the main sequence. By first identifying the positions of the sub-sequence \( T \) and then summing the corresponding SHAP values, we gain a deeper understanding of how specific patterns of actions impact the overall prediction.

\end{enumerate}

An example is provided in Appendix to ensure that the influence of an individual element or a sequence can be aggregated across different contexts within the dataset, providing a method to evaluate the overall impact of specific features in sequence prediction models. Equations \eqref{aggregated1} and \eqref{aggregated2} integrate the SHAP values, which represent marginal contributions of these elements or sequences to the prediction model, presenting a comprehensive explanation of their influence. Following this, the top-k HUSPM \cite{9016187} phase identifies significant patterns in which higher-scoring patterns exhibit greater aggregated SHAP values, offering deeper insights into customer behavior. In the final phase, AL involves continuously refining the model with new data to optimize label efficiency while maintaining high predictive accuracy. Algorithm \ref{algo1} details the process for implementing HUSPM-SHAP, which is used for experimentation in this study. In practical applications, all unlabeled data belong to \( D_{\text{pool}} \). Within Algorithm \ref{algo1}, \( D_{\text{main}} \) corresponds to the previously defined \( D_{\text{initial}} \), and \( D_{\text{pool}} \) is equivalent to \( D_{\text{unlabeled}} \).

\begin{algorithm}[!htbp]
\caption{HUSPM-SHAP for active learning}\label{algo1}
\begin{algorithmic}[1]

\State \textbf{Input:} Initial Dataset \( D \) with Window size \( T \) and Class labels \( L \).
\State \textbf{Output:} Top-k sequences \( \Delta_{k} \).

\State \textbf{I. Data Preparation:}
\State \( X \gets \text{Normalize}(D, T) \). 
\State \( y \gets \text{ExtractLabels}(D, L) \). 
\State \( [X_{\text{train}}, y_{\text{train}}, X_{\text{test}}, y_{\text{test}}] \gets \text{Split}(X, y) \).
\State \( D_{\text{main}} \gets \text{Partition of } D \text{ for initial training and testing} \).
\State \( D_{\text{pool}} \gets \text{Remaining part of } D \text{ for active learning process} \).

\State \textbf{II. Build ML Model:}
\State \( M \gets \text{Model}(\text{Architecture}, \text{Loss}, \text{Optimizer}) \).
\State \( M.\text{fit}(X_{\text{train}}, y_{\text{train}}) \).

\State \textbf{III. Calculate SHAP Values:}
\State \( \text{Let } C \text{ be the classification category based} \)
\Statex \( \text{on model performance \(P \), where:} \)
\State \parbox[t]{\dimexpr\linewidth-\algorithmicindent}{
\( C \gets \begin{cases} 
  C_{\text{positive}}, & \text{if } \text{P}(M, C_{\text{positive}}) < 
  \text{P}(M, C_{\text{negative}}) \\
  C_{\text{negative}}, & \text{otherwise}
\end{cases} \) 
\State \( S \gets \text{Subset}(X_{\text{train}}) \).
\State \( \text{for each } s_i \in S, \text{compute } \phi_{ij} = \text{SHAP}(s_i, j, C) \), where \( j \) is the position in sequence \( s_i \). 
}

\State \textbf{IV. Conduct HUSPM-SHAP:}
\State \( \Delta \gets \text{GenerateSequences}(S) \). 
\State \( \Phi \gets \text{HUSPM}(\Delta, SHAP) \). 
\State \( \text{for each } \tau \in \Phi: U(\tau) \gets \sum_{j} SHAP(s_i, j, C) \). 
\State \( \Delta_{k} \gets \text{Top-k}(\Phi, U) \). 

\State \textbf{V. Active Learning Process and Validation:}
\State \( \mathcal{A} \gets \text{InitStrategy}(\Delta_{k}) \). 
\For{\( i = 1 \) to \( N \)}
    \State \( [D_{\text{pool}}, X_{\text{train}}, y_{\text{train}}] \gets \text{Update}(D_{\text{pool}}, \mathcal{A}, \Delta_{k}) \).
    \State \( M.\text{retrain}(X_{\text{train}}, y_{\text{train}}) \).
    \State \( \text{Evaluate}(M, X_{\text{test}}, y_{\text{test}}) \).
\EndFor
\end{algorithmic}
\end{algorithm}

In Algorithm \ref{algo1}, the symbol \(\Delta\) represents the set of sequences generated during the HUSPM-SHAP process from the input sequences \(S\). The notation \(\Phi\) denotes the set of sequences after applying HUSPM, where SHAP values are incorporated as utility indicators. For each sequence \( s_i \), the SHAP value \( \phi_{ij} \) of each element at position \( j \) is computed by evaluating its marginal contribution to the model's prediction relative to all possible feature coalitions. The aggregated utility \( U(\tau) \) of a sequence \( \tau \in \Phi \) is computed by summing the SHAP values across all elements in the sequence. If a sub-sequence appears multiple times within the same sequence, the maximum SHAP value among its occurrences is used. Similarly, for aggregated utility computed at the sub-sequence level, the same approach is applied, where the highest aggregated SHAP value among repeated occurrences is selected. From \(\Phi\), the top-k sequences with the highest utility, \(\Delta_k\), are selected for labeling. Finally, \(A\) represents the active learning strategy initialized with \(\Delta_k\), guiding the selection of unlabeled data points in subsequent iterations.

\begin{figure}[!htbp]
		\centering
		\centerline{\includegraphics[scale=0.4]{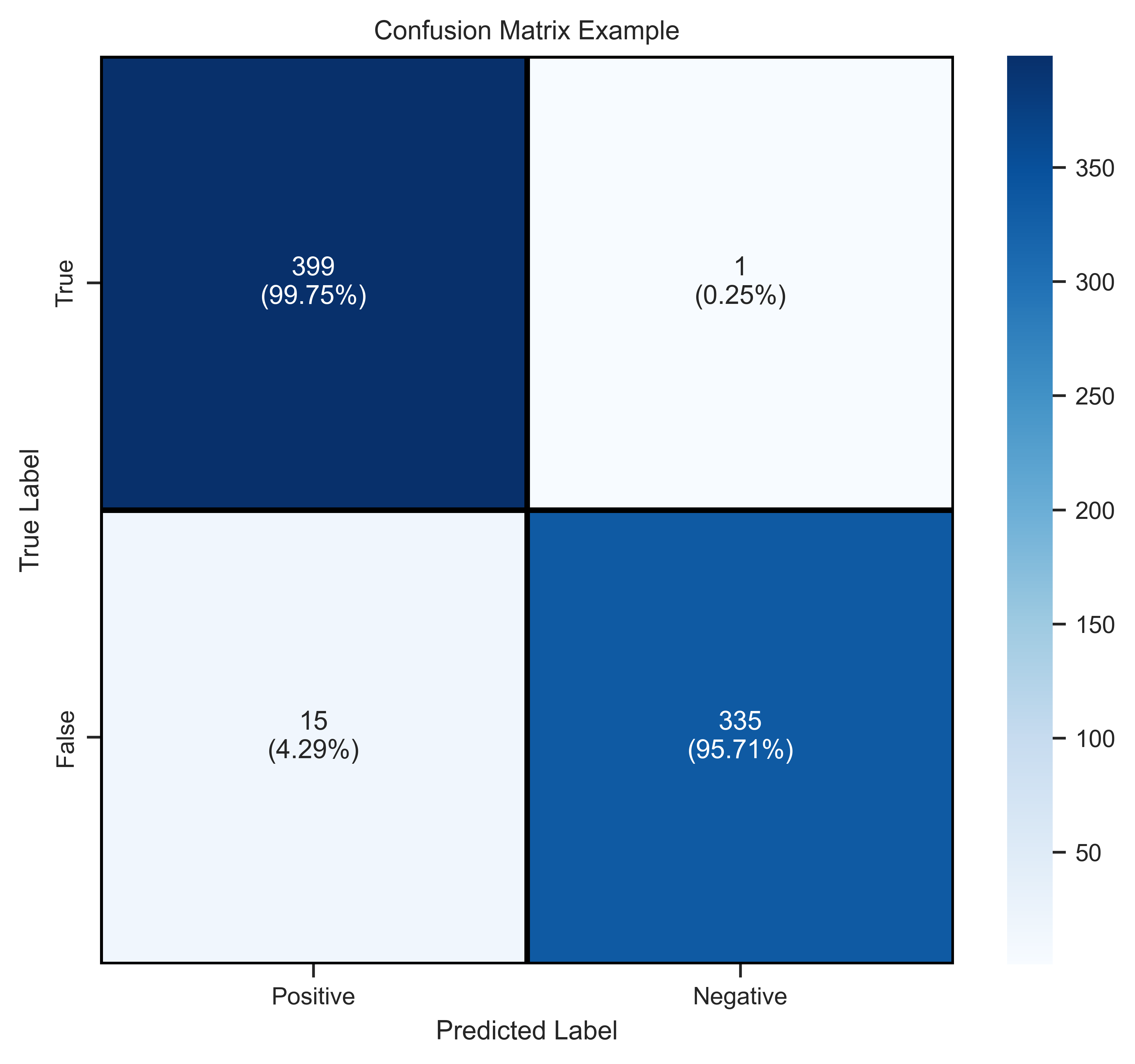}}
		\caption{A confusion matrix example}
        \label{fig4}
	\end{figure}

Starting from a labeled pool, sequences of user actions are fed into a prediction model \(M\), which evaluates SHAP values for sampled instance. The "Build ML Model" step refers to the creation of a general supervised ML model for sequence data. While the experimental validation in this study utilizes LSTM-based models, HUSPM-SHAP framework itself is model-agnostic. That is, it can be applied to any classifier for which SHAP values can be computed, including but not limited to tree-based and neural network models. Since SHAP values are model-agnostic, any sequence-based model can be used in this step depending on the specific application requirements. Before calculating SHAP values, it is necessary to define the positive or negative categories of the SHAP values. Algorithm \ref{algo1} prioritizes the categories of classification that perform poorly and calculates their SHAP values. For example, in the confusion matrix presented in Fig. \ref{fig4}, the success rate of negative class predictions is lower. Therefore, in the calculation of SHAP values, we prioritize selecting parameters for the negative class. As described in Algorithm \ref{algo1}, we first select the negatively performing category to compute sequences of high utility. The underlying logic is that we prioritize the importance of the poorly performing category to guide the inclusion of future new data. It is important to note that the selection of which class (positive or negative) to compute SHAP values for can significantly impact the resulting set of high-utility patterns, as each class may highlight different behavioral sequences. Therefore, the target class should be selected according to the primary analytic objective, such as maximizing purchase detection or minimizing false positives.

\begin{figure*}[!htbp]
		\centering
		\centerline{\includegraphics[scale=0.7]{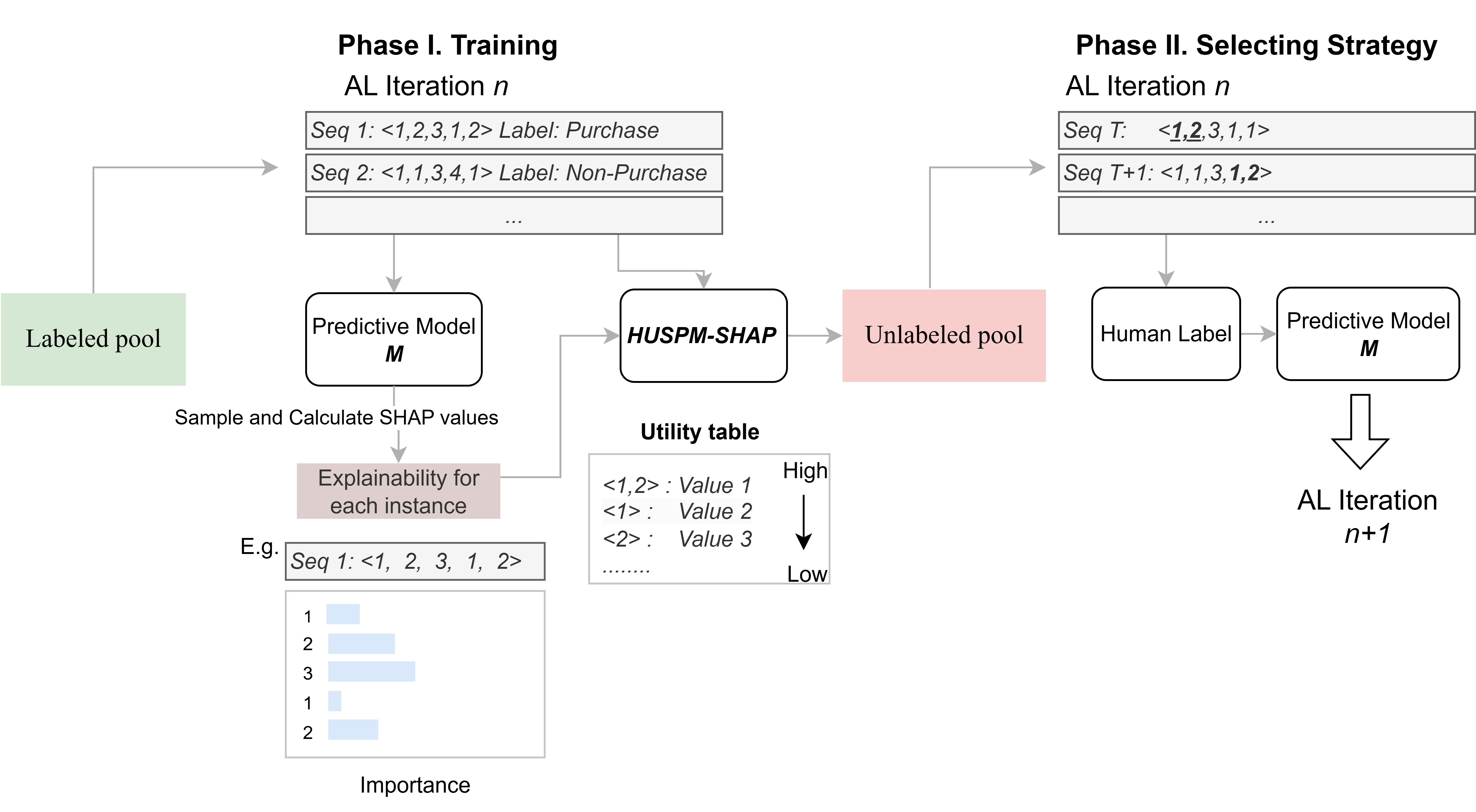}}
		\caption{Graphical illustration of the calculation logic of HUSPM-SHAP in pool-based active learning}
        \label{fig3}
	\end{figure*}

These SHAP values inform the HUSPM-SHAP process, which ranks the utilities of sequences called patterns based on their frequency and summation of SHAP values, and gives priority to those containing mining patterns into an unlabeled pool for further labeling. A human operator adds the necessary labels, after which the sequences are reintegrated into the prediction model \(M\), completing the active learning cycle. This visual representation captures the iterative process of enhancing the model with high-utility sequences, which are expected to improve the model's performance over time. Fig. \ref{fig3} provides a visual elaboration of the HUSPM-SHAP algorithm's workflow.

\section{Experimental Evaluation}
In this section, we first introduce the dataset used in this study and discuss the evaluation measures. At the end of this section, several experiments are then conducted to evaluate the performance of the proposed approach and the compared methods. In this study, we will use precision, accuracy, F1 score, recall rate, and Matthews correlation coefficient (MCC) \cite{chicco2020advantages} as metrics to measure the performance of our model predictions. Especially notable is the F1 Score, which has been extensively utilized in predictions involving clickstream data \cite{requena2020shopper}, \cite{tagliabue2021sigir}.
\subsection{Dataset}
In this study, we consider an e-commerce clickstream datasets named Coveo\footnote{\url{https://github.com/coveooss/SIGIR-ecom-data-challenge}}. The Coveo dataset has over 4 million shopping sessions with detailed user interaction data, including page views, product details, and session events, enriched with metadata like price categories and vectorized product descriptions. This dataset provides resources for e-commerce studies, highlighting different aspects of consumer interactions. In this study, we employ stratified sampling \cite{singh2014sampling} to ensure an equal amount of data in each iteration of experiments on the active learning strategy. This preparatory work allows us to make proper use of the dataset.

\subsection{Baseline Model and Experimental Setup}

In this study, several experiments are conducted to evaluate the effectiveness of active learning techniques in the context of e-commerce clickstream data analysis, specifically targeting the prediction of user browsing behaviors that do result in subsequent purchases or not. Since accurately predicting purchases yields more beneficial outcomes compared to incorrectly predicting non-purchases, we focused on evaluating metrics that reflect true positive outcomes where purchases are correctly predicted.
First, we need to identify several models for evaluating the subsequent active learning assessment. According to Requena et al. \cite{requena2020shopper}, a high-performing model for classification tasks is used as the baseline. This model is named the LSTM discriminative classifier and utilizes the LSTM output at the last step (S2L last). The architecture of the model is shown in Table \ref{model_archi}. 

\begin{table}[!htbp]
\centering
\caption{Baseline Model Architecture}\label{model_archi}
\begin{tabular}{lc}
\hline
\textbf{Layer (Type)} \\ \hline
Embedding (Input dim=200, Output dim=40) \\ 
LSTM (100 units, return\_sequences=True) \\ 
Dropout (rate=0.2) \\ 
Batch Normalization \\ 
LSTM (50 units, return\_sequences=False) \\ 
Dense (2 units, Softmax activation) \\ \hline
\end{tabular}
\end{table}

The baseline model is trained with early-window settings of $5$ and $14$. The reason for choosing early window lengths of 5 and 14 is based on the experimental results from the study by Requena et al \cite{requena2020shopper}. Their study showed that predictions with early window lengths between $5$ and $14$ gradually improved the F1 score, demonstrating the potential for early predictions on this dataset. In this study, we also aligned with the minimum and maximum window settings used in Requena et al.'s study \cite{requena2020shopper}. By randomly selecting $10,000$ data points, a sample size determined to sufficiently represent various imbalanced scenarios with relatively fewer purchase cases. This selection allows us to create imbalanced purchase to purchase ratios of $50\%$, $30\%$, and $10\%$, enabling effective testing of the model's performance under conditions of data category imbalance. TABLE \ref{S2L_coveo} shows the performance of the baseline model in which P represents data with purchases, and NP represents data without purchases.

\begin{table}[htbp!]
\centering
\caption{Performance using S2L Last method with various sample sizes and window lengths on Coveo dataset} \label{S2L_coveo}
\begin{tabular}{lccc}
\hline
\multicolumn{4}{c}{\textbf{S2L Last Model}} \\ \hline
\textbf{}                           & \textbf{10\% P, 90\% NP} & \textbf{30\% P, 70\% NP} & \textbf{50\% P, 50\% NP} \\ \hline
\multicolumn{4}{c}{Coveo dataset: 10,000 points, Early window length: 5} \\ \hline
Precision                  & 97.70\%               & 97.31\%               & 89.61\%               \\
Accuracy                   & 93.57\%               & 85.33\%               & 80.80\%               \\
Recall                     & 95.36\%               & 84.34\%               & 70.72\%               \\
F1                         & 96.51\%               & 90.36\%               & 79.05\%               \\
MCC                        & 55.59\%               & 63.15\%               & 63.26\%               \\ \hline
\multicolumn{4}{c}{Coveo dataset: 10,000 points, Early window length: 14} \\ \hline
Precision                  & 96.38\%               & 92.54\%               & 84.71\%               \\
Accuracy                   & 92.97\%               & 88.93\%               & 87.93\%               \\
Recall                     & 95.92\%               & 91.85\%               & 91.11\%               \\
F1                         & 96.15\%               & 92.20\%               & 87.80\%               \\
MCC                        & 55.36\%               & 73.19\%               & 76.09\%               \\ \hline
\end{tabular}
\end{table}

\subsection{Analysis of Feature Importance via SHAP Values}
\begin{figure}[!htbp]
		\centering
		\centerline{\includegraphics[scale=0.25]{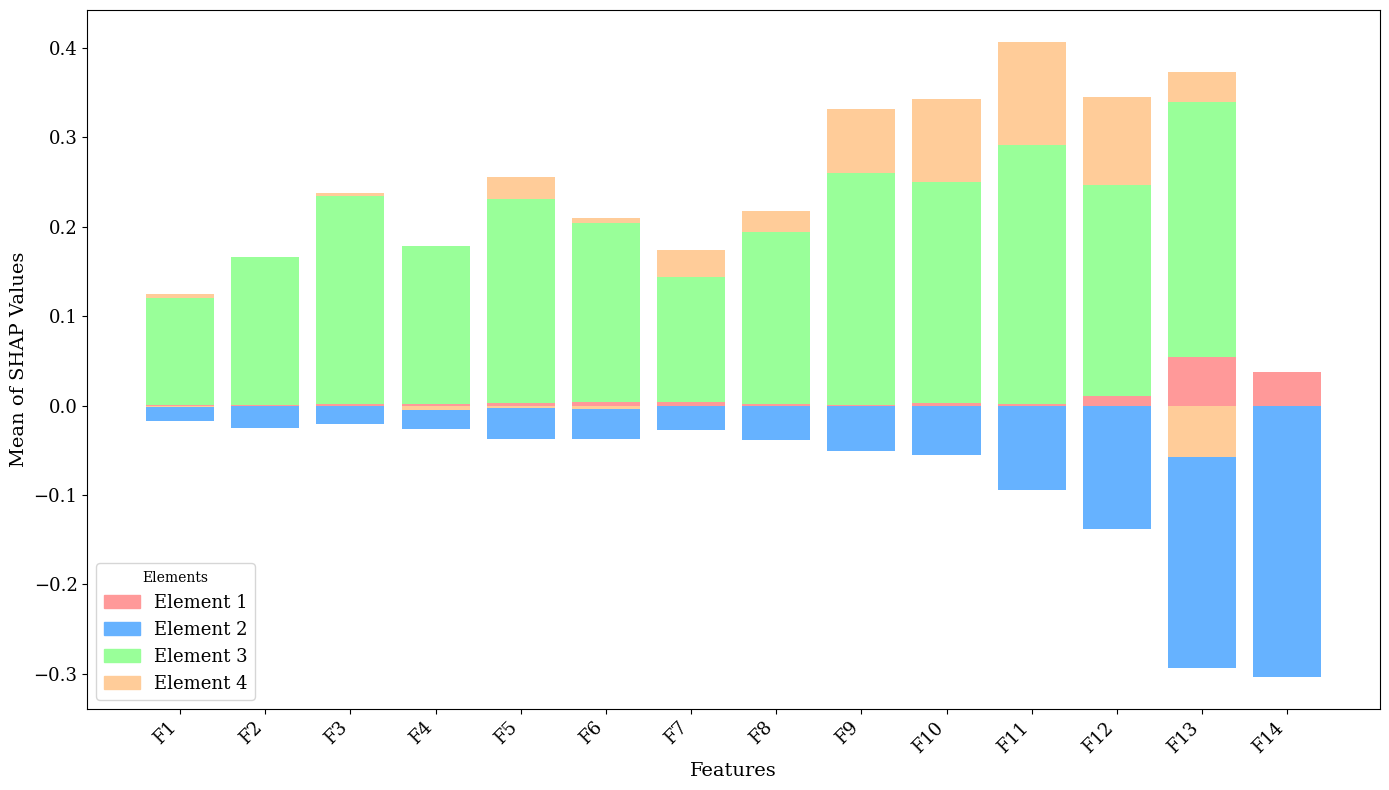}}
		\caption{A stacked bar chart of mean of sampling SHAP values}
		\label{fig5}
	\end{figure}

To better understand the impact of individual features on purchase prediction, we analyze the feature importance using SHAP values derived from our baseline model. Furthermore, the SHAP value analysis demonstrates the model's ability to attribute importance to key features. In the scenario with an early window length of $14$ and a 50\% purchase instances, Fig. \ref{fig5} presents feature importance based on SHAP values from 100 samples with positive and negative values, and illustrates that element $3$ holds the greatest significance in predicting purchase behavior. However, when applying the methodology proposed by Kara et al. \cite{kara2022shap} to identify the maximum SHAP value, the result shows element $1$ instead of intuitive element $3$. From this example, it can show the difference on two different query strategies.

\begin{figure}[!htbp]
	\centering
	\centerline{\includegraphics[scale=0.30]{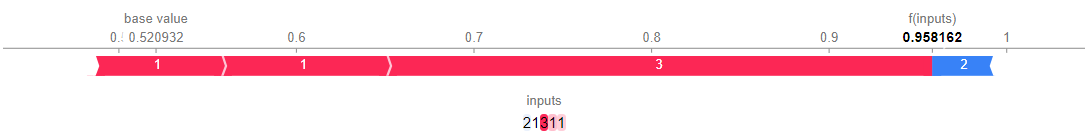}}
	\caption{A SHAP force plot example showing the individual clickstream sequence contribution}
	\label{fig9}
\end{figure}

\begin{figure}[!htbp]
		\centering
		\centerline{\includegraphics[scale=0.25]{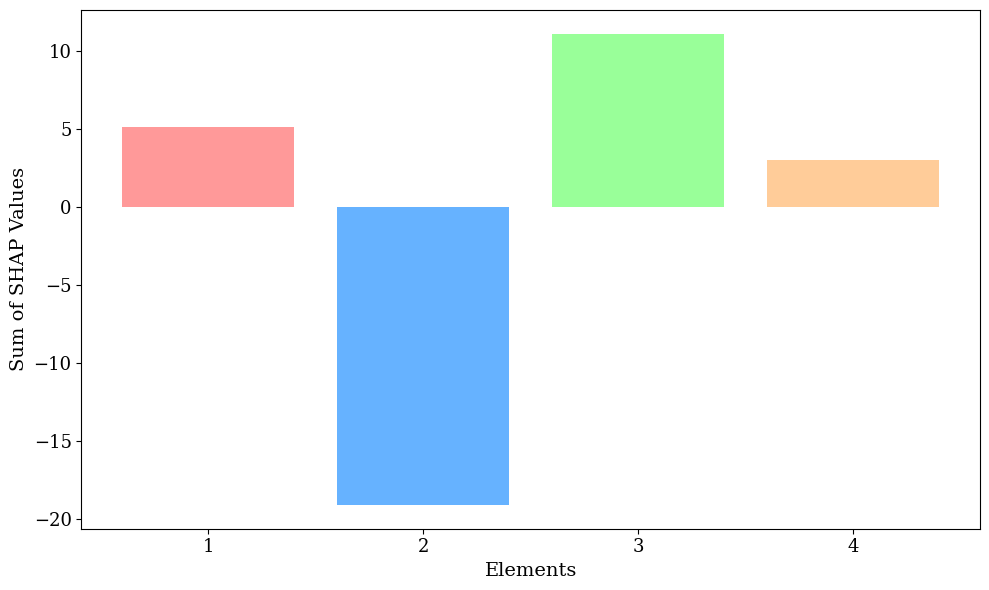}}
		\caption{Sum of sampling SHAP values for each element}
		\label{fig6}
	\end{figure}

In the same scenario illustrated in Fig. \ref{fig5}, we also consider the sum of SHAP values from the dimension of a single element, as shown in the Fig. \ref{fig6}. We observe that element $3$ has a higher contribution to the overall prediction of purchases. Therefore, when we consider a single-element sequence using the HUSPM-SHAP approach, selecting unlabeled data that contains element $3$ would be more beneficial for improving the prediction results and the selection process. Fig. \ref{fig9} shows the contributions of different features to the model's output in a clickstream sequence, where red arrows indicate positive contributions that increase the prediction, and the blue arrow represents a negative contribution that decreases it. It demonstrates that element $3$ has a higher contribution, while element $2$ provides a negative contribution. If the SHAP values' positive and negative class parameters are swapped, the computed utility results would be different, highlighting the sensitivity of utility calculations to the choice of SHAP value assignments.

Using the scenario illustrated in Fig. \ref{fig6} as an example, with the HUSPM-SHAP algorithm from Algorithm \ref{algo1} and Fig. \ref{fig3}, here are the top-5 high-utility subsequences we identified. In this example, we include data instances containing these high-utility sequences in order of their ranking. Here, high-utility subsequences refer to sequences that are either highly frequent, highly valuable, or both, ensuring that the selected data contributes to the active learning process.

\begin{table}[!htbp]
\centering
\caption{Top 5 high utility sequences calculated by HUSPM-SHAP}\label{utilityexample}
\begin{tabular}{lcc}
\hline
Rank & Sequence & Utility by SHAP value \\ \hline
Top 1 & $\langle 3 \rangle$ & 9.68 \\ 
Top 2 & $\langle 1, 3 \rangle$ & 5.81 \\ 
Top 3 & $\langle 1 \rangle$ & 4.36 \\ 
Top 4 & $\langle 2, 3 \rangle$ & 3.87 \\ 
Top 5 & $\langle 2, 1, 3 \rangle$ & 3.80 \\ \hline
\end{tabular}
\end{table}

\subsection{Active Learning Iterative Evaluation and Comparison}

In the experiment, the active learning \cite{settles2009active} phase is designed for six distinct iterations. During each iteration, an additional set of 1,000 data points is selected and added to the training dataset. Crucially, following each increment of 1,000 data points, the S2L last model \cite{requena2020shopper} is retrained from scratch. This process of completely retraining the model after each addition is intended to systematically evaluate the impact of the progressively enlarged training set on the model's prediction accuracy \cite{ash2020warm}. This methodological approach ensures a thorough understanding of how the progressive inclusion of newly labeled data in active learning influences the model's learning effectiveness, particularly in terms of performance improvement and adaptation to previously uncertain instances.

An important aspect of this study is the comparative analysis of different active learning strategies: the HUSPM-SHAP method, random sampling, uncertainty sampling named Uncertainty Sampling - 1st (UNC-1) \cite{sharma2017evidence}, SHAP-based query method \cite{kara2022shap}, Core-set method \cite{sener2017active}, and LLAL method \cite{yoo2019learning}. For UNC-1, we adopt the classical uncertainty sampling baseline. For each iteration, the instance with the smallest difference between the two highest class probabilities is selected, following the margin-based uncertainty principle \cite{lewis1995sequential}. TABLE \ref{tab:model_performance} and TABLE \ref{tab:model_performance2} present the highest value achieved for each metric over the six iterations for every method. After $6$ iterations, the dataset becomes larger than the initial labeled dataset used as the baseline. This ensures that the model has been exposed to a sufficiently large amount of labeled data while maintaining a balance between computational feasibility and experimental validity.

\begin{table}[!htbp]
\small
\setlength{\tabcolsep}{2.2pt}
\centering
\caption{Early Window Length 5: Comparison of model performance across various active learning models (best in each row: bold and underlined).\\
P: Precision, RC: Recall, ACC: Accuracy}
\label{tab:model_performance}
\begin{tabular}{l@{}cccccc@{}}
\hline
\textbf{Metrics} & \textbf{RD} & \textbf{UNC-1} & \textbf{Core-set} & \textbf{LLAL} & \textbf{SHAP} & \textbf{Proposed} \\
\hline
\multicolumn{7}{c}{\textbf{Purchase Ratio: 90\%}} \\
P    & 98.51\% & \textbf{\underline{98.57\%}} & 98.39\% & 97.70\% & 98.39\% & \textbf{\underline{98.57\%}} \\
ACC  & 93.30\% & 93.93\% & \textbf{\underline{94.00\%}} & 93.90\% & 93.93\% & \textbf{\underline{94.00\%}} \\
RC   & 94.27\% & 94.96\% & 95.19\% & \textbf{\underline{95.70\%}} & 95.12\% & 95.03\% \\
F1   & 96.34\% & 96.73\% & 96.76\% & 96.69\% & 96.73\% & \textbf{\underline{96.77\%}} \\
MCC  & \textbf{\underline{58.65\%}} & 56.18\% & 57.30\% & 58.55\% & 34.85\% & 56.80\% \\
\hline
\multicolumn{7}{c}{\textbf{Purchase Ratio: 70\%}} \\
P    & \textbf{\underline{97.99\%}} & 97.69\% & 97.55\% & 97.59\% & 97.73\% & 97.55\% \\
ACC  & 84.97\% & 85.43\% & 85.37\% & 85.30\% & 85.37\% & \textbf{\underline{85.47\%}} \\
RC   & 83.33\% & 84.21\% & 84.23\% & 84.13\% & 84.12\% & \textbf{\underline{84.33\%}} \\
F1   & 90.07\% & 90.45\% & 90.40\% & 90.36\% & 90.42\% & \textbf{\underline{90.46\%}} \\
MCC  & 63.43\% & 63.50\% & 63.29\% & 63.12\% & 63.34\% & \textbf{\underline{63.56\%}} \\
\hline
\multicolumn{7}{c}{\textbf{Purchase Ratio: 50\%}} \\
P    & 75.60\% & 75.54\% & \textbf{\underline{75.86\%}} & 75.73\% & 75.67\% & \textbf{\underline{75.86\%}} \\
ACC  & 77.50\% & 81.60\% & 81.53\% & 81.33\% & 81.57\% & \textbf{\underline{81.67\%}} \\
RC   & 81.60\% & \textbf{\underline{86.84\%}} & 86.43\% & 86.16\% & 86.66\% & 86.69\% \\
F1   & 80.81\% & 80.79\% & 80.80\% & 80.61\% & 80.79\% & \textbf{\underline{80.92\%}} \\
MCC  & 63.85\% & 63.87\% & 63.66\% & 63.24\% & 63.77\% & \textbf{\underline{63.95\%}} \\
\hline
\end{tabular}
\end{table}

\begin{table}[!htbp]
\small
\setlength{\tabcolsep}{2.6pt}
\centering
\caption{Early Window Length 14: Comparison of model performance across various active learning models (best in each row: bold and underlined).}
\label{tab:model_performance2}
\begin{tabular}{l@{}cccccc@{}}
\hline
\textbf{Metrics} & \textbf{RD} & \textbf{UNC-1} & \textbf{Core-set} & \textbf{LLAL} & \textbf{SHAP} & \textbf{Proposed} \\
\hline
\multicolumn{7}{c}{\textbf{Purchase Ratio: 90\%}} \\
P & \textbf{\underline{97.04\%}} & 96.86\% & 96.27\% & 96.97\% & 96.39\% & 96.53\% \\
ACC  & 93.77\% & 93.77\% & \textbf{\underline{93.80\%}} & 93.63\% & 93.67\% & \textbf{\underline{93.80\%}} \\
RC    & 96.16\% & 96.33\% & \textbf{\underline{96.91\%}} & 96.09\% & 96.67\% & 96.67\% \\
F1        & \textbf{\underline{96.60\%}} & 96.59\% & 96.59\% & 96.52\% & 96.53\% & \textbf{\underline{96.60\%}} \\
MCC       & 59.66\% & 60.31\% & \textbf{\underline{62.69\%}} & 58.80\% & 60.14\% & 61.24\% \\
\hline
\multicolumn{7}{c}{\textbf{Purchase Ratio: 70\%}} \\
P & 94.24\% & \textbf{\underline{94.76\%}} & 93.49\% & 93.87\% & 93.63\% & 91.51\% \\
ACC  & 89.17\% & 88.80\% & 89.13\% & 89.30\% & 89.23\% & \textbf{\underline{89.37\%}} \\
RC    & 90.77\% & 89.92\% & 91.33\% & 90.86\% & 91.34\% & \textbf{\underline{93.63\%}} \\
F1        & 92.48\% & 92.28\% & 92.40\% & 92.36\% & 92.47\% & \textbf{\underline{92.56\%}} \\
MCC       & 73.33\% & 72.27\% & 73.43\% & 73.06\% & 73.65\% & \textbf{\underline{74.00\%}} \\
\hline
\multicolumn{7}{c}{\textbf{Purchase Ratio: 50\%}} \\
P & 89.00\% & \textbf{\underline{94.01\%}} & 88.29\% & 90.24\% & 88.61\% & 89.26\% \\
ACC  & 88.20\% & 87.30\% & 88.10\% & 87.83\% & 87.87\% & \textbf{\underline{88.23\%}} \\
RC    & 88.09\% & 83.33\% & 88.46\% & 86.58\% & 87.81\% & \textbf{\underline{87.95\%}} \\
F1        & 88.54\% & 88.35\% & 88.38\% & 88.37\% & 88.21\% & \textbf{\underline{88.60\%}} \\
MCC       & 76.38\% & 75.16\% & 76.19\% & 75.70\% & 75.72\% & \textbf{\underline{76.45\%}} \\
\hline
\end{tabular}
\end{table}

\begin{figure*}[!htbp]
		\centering
		\centerline{\includegraphics[scale=0.40]{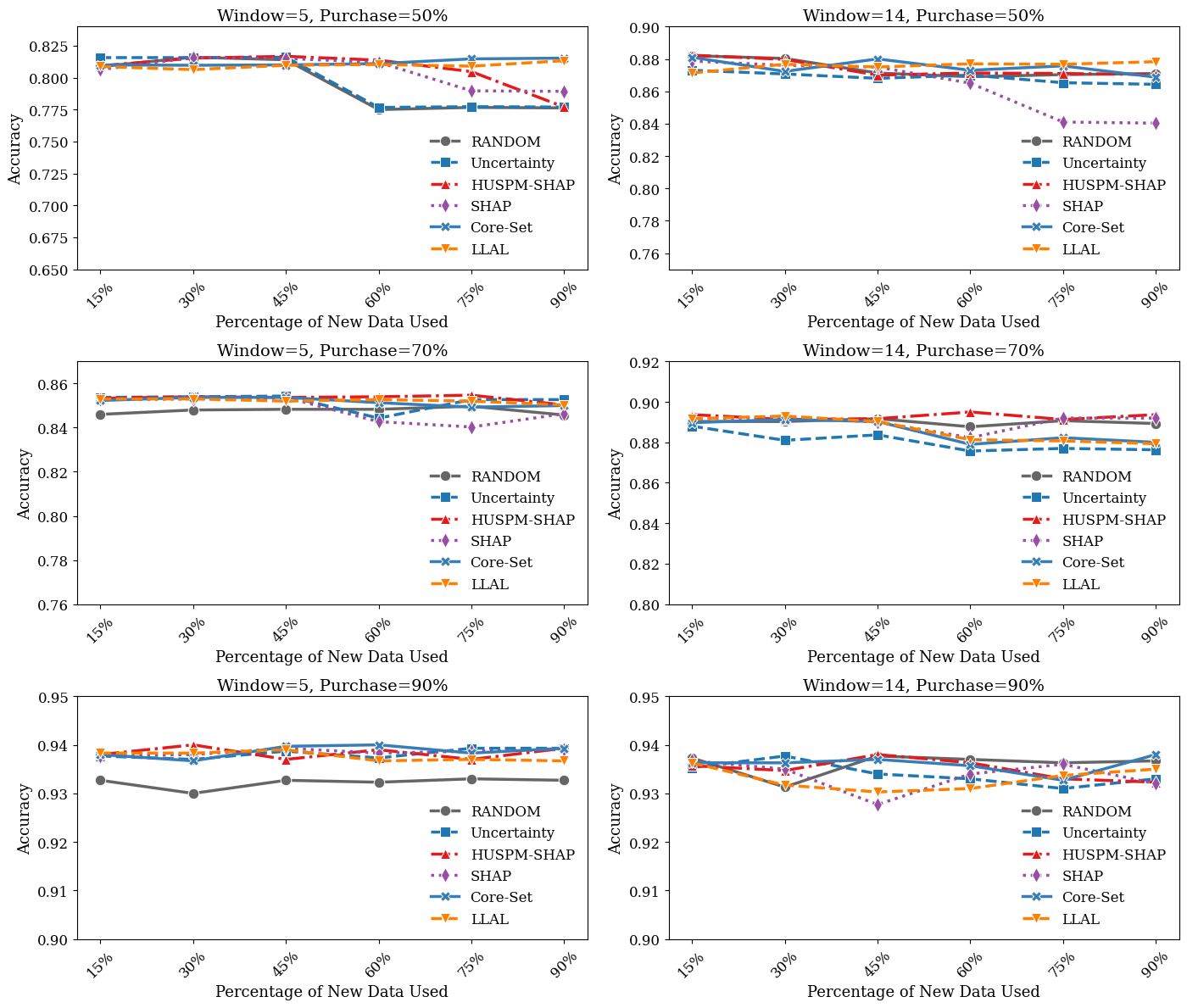}}
		\caption{Comparative performance of active learning strategies across data utilization levels}
		\label{fig2}
	\end{figure*}

In the comparative analysis of active learning models across varying early-window settings, a clear trend emerges: the HUSPM-SHAP model displays robust performance, highlighted by its consistent accuracy and F1 scores, especially in experiments where the purchase-to-non-purchase ratio is set to $10\%$:$90\%$ and $30\%$:$70\%$, representing different levels of data imbalance. This trend is consistent across window settings of 5 and 14, illustrating that HUSPM-SHAP utilizes both relatively long and short sequences to predict browsing behavior.

\subsection{Abliation Study}
To assess the contribution of each component in our framework, we conducted ablation experiments under both balanced ($50\%$:$50\%$) and imbalanced ($10\%$:$90\%$) scenarios with a window length of 14. We compared four settings: random sampling (RD), the SHAP-based query method (SHAP), sequential pattern mining without SHAP values (SPM), and our HUSPM-SHAP framework. For HUSPM-SHAP, we implemented two backbone models, LSTM and Transformer (TR) \cite{vaswani2017attention}, to examine the impact of model architecture. All methods were evaluated on identical data splits and active learning procedures using standard classification metrics.

\begin{table}[!htbp]
\small
\centering
\caption{Ablation study: Performance comparison of different backbones for HUSPM-SHAP (Sequence Length = 15, Purchase Ratio = 10\%).}
\begin{tabular}{lccccc}
\hline
 &  &  &  &\multicolumn{2}{c}{\textbf{HUSPM-SHAP}} \\
\cmidrule(lr){5-6}
\textbf{Metrics} & \textbf{RD} & \textbf{SHAP} & \textbf{SPM} & \textbf{LSTM} & \textbf{TR} \\
\hline
Precision & 89.00\% & 88.61\% & 88.68\% & \textbf{\underline{89.26\%}} & 88.35\% \\
Accuracy  & 88.20\% & 87.87\% & 88.10\% & \textbf{\underline{88.23\%}} & 87.03\% \\
Recall    & 88.09\% & 87.81\% & \textbf{\underline{88.16\%}} & 87.95\% & 86.61\% \\
F1        & 88.54\% & 88.21\% & 88.42\% & \textbf{\underline{88.60\% }}& 87.47\% \\
\hline
\end{tabular}
\end{table}

\begin{table}[!htbp]
\small
\centering
\caption{Ablation study: Performance comparison of HUSPM-SHAP using different backbone models (Sequence Length = 15, Purchase Ratio = 50\%).}
\begin{tabular}{lccccc}
\hline
 &  &  &  &\multicolumn{2}{c}{\textbf{HUSPM-SHAP}} \\
\cmidrule(lr){5-6}
\textbf{Metrics} & \textbf{RD} & \textbf{SHAP} & \textbf{SPM} & \textbf{LSTM} & \textbf{TR} \\
\hline
P & 97.04\% & 96.39\% & 96.23\% & 96.53\% & \textbf{\underline{97.40\%}} \\
Acc  & 93.77\% & 93.67\% & 93.67\% & \textbf{\underline{93.80\%}} & 93.37\% \\
RC    & 96.16\% & 96.67\% & \textbf{\underline{96.80\%}} & 96.67\% & 95.42\% \\
F1        & \textbf{\underline{96.60\%}} & 96.53\% & 96.52\% & \textbf{\underline{96.60\%}} & 96.40\% \\
\hline
\end{tabular}
\end{table}
The ablation results indicate that HUSPM-SHAP, when paired with an LSTM backbone, achieves robust and often superior predictive performance across both imbalanced and balanced purchase scenarios (e.g., 10\% and 50\%). SHAP-based and random sampling baselines serve as important references, with random sampling providing a lower bound and SHAP offering an interpretable upper bound among non-sequential methods. When the backbone is changed, the observed performance drop demonstrates that the effectiveness of sequential pattern mining and active learning not only relies on the data mining strategy but is also critically influenced by the suitability of the underlying model architecture. This is especially evident in imbalanced settings, where LSTM, SPM, and even random sampling maintain relatively stable F1 scores and accuracy, highlighting resilience in minority class prediction. To sum up, these findings emphasize that optimal sequence-based active learning performance can only be achieved when both the mining strategy and model backbone are properly matched. It is insufficient for optimizing only one component.

\subsection{Summary of Key Findings}
Upon analyzing active learning models under class imbalance conditions where purchase instances constitute $10\%$, $30\%$, and $50\%$ of the dataset, HUSPM-SHAP consistently delivers high performance metrics, e.g., accuracy, recall, F1 scores, and MCC, demonstrating its effectiveness and robustness in managing different levels of class distribution.

\begin{figure}[!htbp]
		\centering
		\centerline{\includegraphics[scale=0.35]{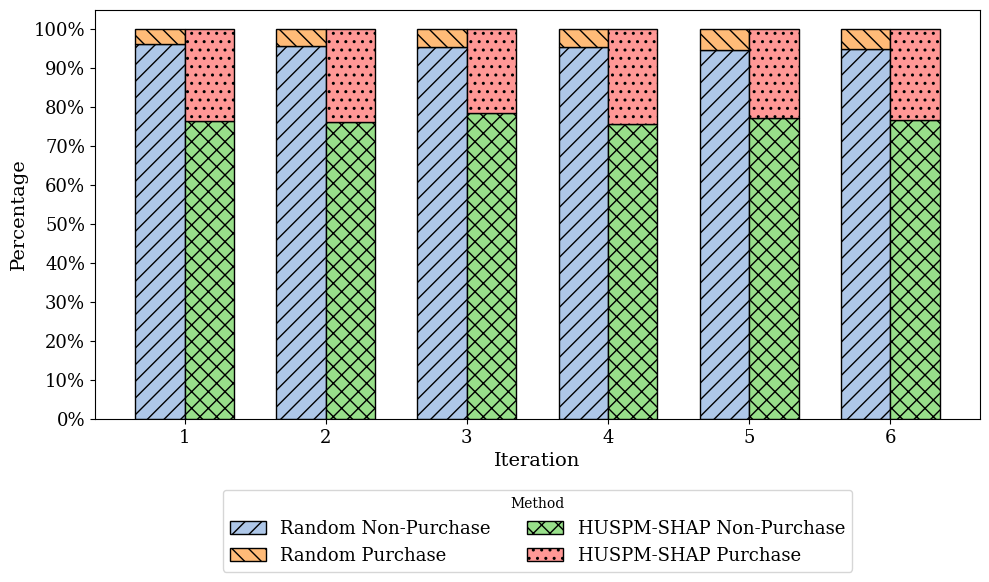}}
		\caption{Comparison of purchase and non-purchase percentage between random and HUSPM-SHAP methods using 10\% purchase data}
		\label{fig7}
	\end{figure}

An interesting observation in the experiments is that in imbalanced datasets, using the example of a $10\%$ purchase case with an early window length of $14$, the data extracted using HUSPM-SHAP, has a higher chance of including purchase data, as shown in Fig. \ref{fig7}.

As the class balance shifts towards a $50\%$–$50\%$ split, HUSPM-SHAP continues to demonstrate robust performance, achieving the highest accuracy and F1 scores among all tested methods. While random sampling and uncertainty sampling yield competitive results for certain metrics and scenarios, their performance is generally comparable but less consistent across all settings. The Core-set and LLAL approaches also perform well in specific cases and class ratios. The SHAP-based query model \cite{kara2022shap} maintains strong recall in imbalanced cases, making it valuable when detecting the dominant class is essential. HUSPM-SHAP is distinguished by its overall stability and consistently strong results across key metrics including accuracy, recall, F1 score, and MCC. This is especially apparent for imbalanced e-commerce datasets where non-purchase events predominate. While some baseline models yield higher performance for certain metrics, they lack interpretability and transparency in the sampling process. In contrast, HUSPM-SHAP provides both competitive accuracy and model explainability. As illustrated in Fig.~\ref{fig7}, HUSPM-SHAP is also effective at selecting purchase-related instances from the unlabeled data pool, thereby improving minority class representation. Overall, HUSPM-SHAP demonstrates adaptability and reliability across class distributions, highlighting its suitability for e-commerce applications.

Fig.~\ref{fig2} provides a comparative view of active learning strategies across different levels of data utilization, illustrating that HUSPM-SHAP maintains stable and accurate performance throughout the iterative labeling process. This highlights HUSPM-SHAP's adaptability and robustness under different data availability conditions. For code implementation details, please refer to the provided GitHub\footnote{Github: \url{https://github.com/dannyycwang/HUSPM-SHAP/tree/main}}.

In summary, this study confirms the impact of class imbalance observed by Kara et al. \cite{kara2022shap}. Our findings introduce a distinction regarding the role of SHAP value class selection. We found that explicitly distinguishing between positive and negative classes in SHAP value analysis is pivotal for generating top-k sequences, which yields totally distinct outcomes. The differences observed can be explained by the methodological approaches: Kara et al.'s study \cite{kara2022shap} selects the maximal value from the ensemble of SHAP values, whereas this study adopts a comprehensive aggregation of the full range of SHAP values. Consequently, the mean values for positive and negative instances remain distinct.

\section{Conclusion}

In conclusion, this study presents the value of active learning in clickstream data analysis, particularly in its application to the HUSPM-SHAP model. Unlike common SHAP computation, which focuses on the evaluation of individual elements, HUSPM-SHAP integrates SHAP values with HUSPM to find high-utility sequential patterns that enhance the performance of the prediction model. By focusing on the explainable data points, the proposed HUSPM-SHAP demonstrates not only increased efficiency in labeling but also maintained accuracy in model training. This approach is particularly effective in unraveling the complexities of non-converting visitors, providing key insights for e-commerce platforms to refine their strategies and better cater to potential customer preferences. The model's consistent performance across different scenarios (e.g., achieving comparable or slightly better results in all tested settings), along with its ability to provide interpretable decision-making, demonstrates its value for predictive modeling in e-commerce applications. To the best of our knowledge, this is the first attempt to use high utility mining concepts within an active learning framework, potentially opening up new opportunities for future explorations in this domain. This shift towards predictive analysis offers a new way to address challenges like data scale and class imbalance.

\section{Acknowledgment}

To improve readability and language quality, parts of this paper were grammatically revised with the assistance of ChatGPT (OpenAI) \cite{openai_chatgpt}.

\bibliographystyle{IEEEtran}
\bibliography{ref.bib}

\end{document}